\documentclass{ifacconf}
\usepackage{times}
\usepackage{balance}
\usepackage{graphicx}
\usepackage{amsmath} 		
\usepackage{amssymb}  	
\usepackage{dsfont}
\usepackage{mathrsfs}
\usepackage{savesym}
\savesymbol{AND}
\savesymbol{OR}
\savesymbol{NOT}
\savesymbol{TO}
\savesymbol{COMMENT}
\savesymbol{BODY}
\savesymbol{IF}
\savesymbol{ELSE}
\savesymbol{ELSIF}
\savesymbol{FOR}
\savesymbol{WHILE}
\usepackage{algorithmic}
\usepackage{algorithm}
\usepackage[usenames,dvipsnames]{color}
\usepackage[latin1]{inputenc}
\usepackage{subfig}
\usepackage{forloop}
\usepackage{ifthen}
\usepackage{intcalc}
\usepackage{tikz}
\usepackage{textcomp}
\usepackage{float}

\newcommand {\mibf}[1] {\boldsymbol{#1}}

\newcommand {\emilista} {\end{list}}
\newsavebox{\ieeealgbox}

\usepackage{natbib}        
\begin{document}
\begin{frontmatter}

\title{ Physics-based Motion Planning with Temporal Logic Specifications} 

\thanks[footnoteinfo]{This work was partially supported by the Spanish Government through the projects \mbox{DPI2013-40882-P},  \mbox{DPI2014-57757-R} and \mbox{DPI2016-80077-R}. Muhayyuddin is 
supported by the Generalitat de 
Catalunya through the grant FI-DGR 2014.  A.~Akbari is supported by the Spanish Gov. through the grant FPI 2015.}

\author{Muhayyuddin, } 
\author{Aliakbar~Akbari, } 
\author{Jan~Rosell}
\address{Institute of Industrial and Control Engineering (IOC), Universitat Polit\`ecnica de Catalunya (UPC) -- Barcelona Tech, Spain. }

\begin{abstract}
One of the main foci of robotics is nowadays centered in providing a great degree of autonomy to robots. A fundamental step in this direction is to give them the ability to plan in discrete and 
continuous spaces to find the required motions to complete a complex task.
In this line, some recent approaches describe tasks with Linear Temporal Logic (LTL) and reason on discrete actions
 to guide sampling-based motion planning, with the aim of finding dynamically-feasible motions that satisfy the temporal-logic task specifications. The present paper proposes an LTL planning approach 
enhanced with the use of ontologies to describe and reason about the task, on the one hand, and that includes physics-based motion planning to allow the purposeful manipulation of objects, on the 
other hand.  The proposal has been implemented and is illustrated with  didactic examples with a mobile robot in  simple scenarios where some of the goals are occupied with objects that must be 
removed in 
order to fulfill the task.
\end{abstract}


\end{frontmatter}
\section{Introduction}
The field of motion planning is evolving rapidly and one of the main directions is towards enabling robots to perform complex tasks in realistic environments. In this regard, on the one hand, it is 
evolving towards the simultaneous consideration of kinodynamic and physics-based  constraints, like the physics-based motion planning approaches that allow to also consider the purposeful manipulation 
of objects.  On other hand, it is evolving towards the integration with task planning, like the LTL-based approaches that define multiple goal tasks in terms of temporal logic, such as \textit{visit 
region A followed by region C and avoid region B}.

LTL-based motion planning is a hybrid approach of discrete (high-level) and continuous (low-level) planning, that computes the robot trajectories in such a way that they satisfy the temporal 
constraints, represented in terms of an LTL formula. 
To our best knowledge, all state-of-the-art LTL motion planers, such as \cite{fainekos2005,bhatia2010,plaku2012,Lahijanian2012,plaku2013,edelkamp2014},  evaluate the validity of the formula 
using model checking techniques (automaton construction) and in case of feasible formula, determine the collision-free trajectory that satisfy the formula. These planners neither analyze the 
formula against the capability of the robot, nor incorporate  manipulation actions (i.e. they do not consider the dynamic interactions 
between rigid bodies). Therefore they are not able to compute the plan if, for instance,  no collision free trajectory exists for moving between regions that 
the robot must visit, although with the removal of few objects a trajectory could be found.

The straight extension of these approaches in order to handle  manipulation actions is possible, but due to the  high complexity of physics-based motion planning (large search space and highly 
constraint solution set) along with the incorporation of temporal constraints, may lead to a computationally nontractable problem, raising the question about its decidability. Therefore, an efficient 
and powerful framework is required that has the capacity to handle both the temporal goals and the physics-based constraints, along with the purposeful manipulation of the objects if necessary. This 
paper tries to contribute in this line by enhancing the framework introduced in~\cite{Muhayyuddin2015}, that combined the use of ontologies with physics-based motion planning, by allowing now the 
consideration of temporal constraints.

\textit{Contributions.} The main contributions of this paper are: \textit{(1)} the integration of the Linear Temporal Logic within the framework of ontological physics-based motion planning, thus 
allowing the purposeful manipulation of objects, like the execution of push actions to clear regions possibly occupied by objects; \textit{(2)} the proposal of an LTL feasibility evaluation and 
simplification process, that uses knowledge-based reasoning  to evaluate whether the properties of the objects and the robot make the task described by the LTL formula feasible or not and, 
if required, simplifies the formula by skipping the non-valid propositions defined with disjunction relations.

The rest of the paper is structured as follows. Sec.~\ref{sec:rWork} presents 
some relevant work related to physics-based motion planning and LTL-based motion planing. Sec.~\ref{sec:PS} describes the modeling of the world and problem statement. 
Then, Sec.~\ref{sec:approach} explains the framework, the reasoning process, and the planning process.  Implementation issues and some simulation results are explained in 
Sec.~\ref{sec:rnd}. Finally, the conclusions are presented in Sec.~\ref{sec:concl}.
\section{Related Work}\label{sec:rWork}
\subsection{Physics-based Motion Planning}
Motion planning problems focus on computing a collision-free trajectory from a given start state to the goal state in the configuration space. The configuration space~$\mathcal{C}$ represents the set 
of all possible configurations of the robot; it is divided into $\mathcal{C}_{free}$ (free regions) and~$\mathcal{C}_{obs}$ (forbidden regions). To take into account  differential (dynamic) 
constraints the planning is performed in a higher dimensional state space $\mathcal{S}$ that records the system dynamics. For any configuration $q\in \mathcal{C}$ the state of the system is 
represented as $s=(q,\dot{q})$. Planning will be performed in~$\mathcal{S}$ in a similar way as it is done in $\mathcal{C}$ for pure geometric problems.

Kinodynamic motion planning is centered in computing the collision-free trajectories that satisfy the kinodynamic constraints (such as joint limits, bounds on the forces and acceleration).
Sampling-based motion planners~\citep{tsianos2007}, particularly those based on tree data structures, are well suited for this purpose because the state propagation used to incrementally grow the 
data structures can easily take into account the constraints. Moreover, if a dynamic engine is used as state propagator then
 physics-based constraints (gravity and friction) can also be easily incorporated~\citep{sucanK2012},~\citep{plaku202}. Therefore, physics-based motion planning can be considered as an evolved form 
of kinodynamic planning  in which the robot motions also satisfy physics-based constraints.
This, moreover, allows to evaluate the dynamic interaction between rigid bodies and, besides considering only collision-free trajectories, allows the consideration of  manipulation actions (such as 
the push action), thus broadening the range of tasks that can be solved.

The physics-based motion planners rely on sampling-based kinodynamic motion planners, such as Rapidly-Exploring Random Trees (RRT)~\citep{lavalle2001}, Kinodynamic Motion
Planning by Interior-Exterior Cell Exploration (KPIECE)~\citep{sucanK2012}, Synergistic Combination of
Layers of Planning (SyCLoP)~\citep{Plaku2010}, for sampling the states and 
constructing the solution path, while the state propagation is performed using a dynamic engine such as Open Dynamic Engine (ODE)~\citep{OpenDE2007}. The large search space and the evaluation of the 
dynamical interactions 
make physics-based motion planning computationally intensive. A few approaches have been proposed that try to overcome these issues, such as the work of~\cite{zickler2009} that  proposed a
nondeterministic tactic based on a finite state machine to guide the motion planner,  along with the use of action skills to control the sampling.
In a similar direction, the ontological physics-based motion planning approach of \cite{Muhayyuddin2015} performs a knowledge-based reasoning process to compute the way of manipulation for objects, 
thus 
reducing the planning search space.
This knowledge-based framework can be used together with any sampling-based kinodynamic motion planner (such as RRT, KPIECE, SyCLoP), and ODE is set as state propagator.Moreover, this 
framework is also used for computing the motion plan and dynamic cost in integrated task and motion planning approaches such as~\cite{ali2015,ali2016}. The present proposal extends 
this approach for temporal goals described by an LTL formula.

\subsection{LTL-based Motion Planning}
LTL motion planning is a hybrid approach that provides a framework to describe complex motion planning tasks in terms of temporal goals, and that
plans in discrete and continuous spaces. The planning is performed in three steps (1) \textit{Workspace decomposition:} decomposes the robot workspace (using for example a triangular decomposition) 
to 
construct the finite state model of robot's motion; (2) \textit{High level planning:} constructs the discrete plan over the product space of the decomposed workspace and the automaton (that is 
constructed to check the LTL formula) in such a way that the discrete plan satisfies the LTL formula $\phi$; (3) \textit{Low level planning:} implements the high level plan at low level in such a way 
that it also satisfies $\phi$.

LTL-based motion planning approaches are broadly divided into two main categories: controller-based and sampling-based LTL motion planers. The former compute the discrete plan over the decomposed 
workspace and then the controller looks for the dynamically-feasible and collision-free trajectory for each action~\citep{fainekos2009}. The latter consider the integration of the task and motion 
planning steps, proposing a probabilistic search over the hybrid space of discrete and continuous components \citep{bhatia2010, bhatia2011, maly2013, plaku2013, he2015}. The discrete component is 
represented as the 
product space of decomposed workspace and the automaton that satisfies the LTL formula $\phi$, whereas the continuous layer consists of a sampling-based dynamic motion planner that is guided by the 
discrete layer. All these approaches always seek for a collision-free trajectory. Although
some approaches such as \citep{mcmahon2014} incorporate a dynamic engine within an LTL framework, it is used only to consider the robot dynamics and the physics-based constraints, i.e. no
dynamic interactions between rigid bodies are considered while planning.

\section{Preliminaries} \label{sec:PS}

\subsection{Modeling}\label{subsec:modeling}
Consider the world is composed of a set of rigid bodies $\mathcal{B}$, that are categorized into fixed and movable. The former remains fixed throughout the planning process and is represented as 
$\mathcal{B}_{\textrm{fixed}}$, whereas the latter can be moved (pushed) by the robot. The movable bodies are further divided into:
\begin{itemize}
 \item Freely-movable bodies $\mathcal{B}_{\textrm{free}}$: Bodies that can be manipulated freely from any direction.
 \item Constraint-oriented movable bodies $\mathcal{B}_{\textrm{co}}$: Bodies that must be manipulated from certain directions, such as car-like bodies that can only be pushed in the forward or 
backward directions. 
\end{itemize}
The manipulation constraints of a $\mathcal{B}_{\textrm{co}}$ are modeled by defining some part of the body that the robot is allowed to touch, and an associated region, called  
manipulation region. 
(\textit{mRegion}), where the robot must be located in order to interact with the body.
All bodies in the environments can be represented as; 
$\mathcal{B}=\mathcal{B}_{\textrm{free}} \cup \mathcal{B}_{\textrm{co}} \cup \mathcal{B}_{\textrm{fixed}}$.

To store the above stated information and to update it while planning, the knowledge is represented in two levels, the abstract knowledge $\mathcal{K}$ and the instantiated 
knowledge~$\kappa$.
The abstract knowledge is represented using ontologies encoded with the Web Ontology Language (OWL)~\citep{owl2004}. It contains the type of the objects (such as fixed and manipulatable), their 
properties (such as masses and friction coefficients), associated manipulation constraints, kinodynamic properties of the robot (such as joint limits, bounds on the forces, torques and velocities), 
and the LTL operators. The abstract knowledge~$\mathcal{K}$ remains fixed throughout the planning process. The instantiated knowledge~$\kappa$ is the dynamic knowledge, inferred from $\mathcal{K}$ 
through the reasoning process, and updated continuously at each time step. It contains the manipulation constraints that are valid at each particular 
instance of time.

Let $\mathcal{X}$ be the state space of all the bodies in the environment. At any time $t$ a state 
$x\in \mathcal{X}$ is represented as $x(t)=\{s_1 \dots s_k\}$ where $s_i$ represents the position and orientation of $i$-th object in the environment. The instantiated knowledge $\kappa_t^x$ is 
associated to state $x$.
\subsection{Robot Model}
Consider a mobile robot $\mathcal{R}$, and let $\mathcal{S}$ be its state space containing all possible states of the robot. A state $s \in \mathcal{S}$ is represented 
as 
$s=\{p,o,v,w\}$ where 
$p,o,v,$ and $w$ are the position, orientation, linear velocity and angular velocity respectively. The instantiated knowledge $\kappa_t^s$ is associated to $s$. The state of the environment can be 
represented 
as $E=\mathcal{X}\times\mathcal{S}$. 

A trajectory  of the robot is defined by the robot dynamics in the results of control inputs, that are applied for a small time duration $\Delta t$. It can be written as 
\hbox{$s_\textrm{new}=$ \sf \small PROPAGATOR$(s,u,\Delta t)$}, where $u \in \mathcal{U}$ is a control input from the control space $\mathcal{U}$  containing the set of all possible control inputs 
that can be applied to the 
robot. The \hbox{\sf \small PROPAGATOR} will generate a trajectory between state $s$ and $s_\textrm{new}$.  An entire trajectory (\textit{Traj}) of the robot is obtained by applying the 
control inputs 
(starting from start state) repeatedly for small time durations. During the execution of the motion,  
if the robot interact 
with $\mathcal{B}_\textrm{free}$ or $\mathcal{B}_\textrm{co}$ the resulting motion will change the state of the bodies. It implies that control inputs are responsible for updating the state of the 
robot as well as the state of the bodies. Generally the transition function \hbox{\sf \small PROPAGATOR} can be written as \hbox{\sf \small 
PROPAGATOR: $E_i\times\mathcal{U}\rightarrow E_{i+1}$}. The 
instantiated 
knowledge for $E$ can be defined as \hbox{$\kappa=\kappa^s\cup\kappa^x$}.
validity of newly generated state is evaluated by a function \hbox{\sf \small VALIDITYCHECKER: $E\times\kappa\rightarrow\{\top,\bot\}$}. It returns $\top$ iff the newly 
generated state satisfies all the 
constraints imposed by $\kappa$. For each new state of $E$, $\kappa$ is updated by the inference process \hbox{\sf \small INFERENCE: $E_{i+1} \times \kappa_i \rightarrow \kappa_{i+1}$}.
\subsection{Linear Temporal Logic}
Linear temporal logic is a formalism  used to specify tasks by combining propositions with logical and temporal operators. The combination is called an LTL formula $\phi$ \citep{clarke1999}, i.e.
an LTL formula $\phi$ is defined by integrating  propositions with the logic operators \textit{negation} ($\neg$), \textit{conjunction} ($\wedge$), \textit{disjunction} ($\vee$), \textit{equivalence} 
($\Leftrightarrow$), and \textit{implication} ($\Rightarrow$) along with the temporal operators \textit{next} ($\bigcirc$), \textit{always} ($\Box$), \textit{until} ($\sqcup$), and 
\textit{eventually} 
($\diamondsuit$).

Let $\Pi$ be the set of atomic propositions, \hbox{$\Pi=\{\pi_1,\dots,\pi_n\}$}, each $\pi_i$ representing a statement such as \textit{``robot is in region $P_i$"}. Every $\pi_i \in \Pi$ is 
a 
formula, and if $\phi$ and $\psi$ are formulas, then new formulas can be defined  using the following grammar:
\begin{center}
$\neg\phi$,\; $\phi\wedge\psi$,\; $\phi\vee\psi$,\; $\diamondsuit\phi$,\; $\phi\sqcup\psi$,\; $\phi\bigcirc\psi$
\end{center}
 As an example, the formula to visit the regions $P_1,\;P_2,\;P_3$ in an ordered way can be represented as $\phi = \diamondsuit (\pi_1 \wedge \diamondsuit ( \pi_2 \wedge 
\diamondsuit(\pi_3 ))) $

The semantics of LTL formula are defined over infinite traces of a system. Let $\sigma=\tau_0,\tau_1,\dots\tau_{\infty}$  represent the infinite trace with $\tau_i\in2^\Pi$, and $\sigma \models \phi$ 
represent $\sigma$ satisfies $\phi$ iff there exists a finite prefix $\sigma_i=\tau_0,\tau_1,\dots,\tau_{i-1}$ of $\sigma$ that satisfies $\phi$. 
 
Syntactically co-safe formulas are a special class of LTL formulas.  When written in positive normal form (i.e. when the negation operator occurs only in front of atomic 
propositions), they only contain the eventually, next and until operators. They can be interpreted over a finite trace and their validity can be checked using nondeterministic finite 
automata~\citep{kupferman2001}.
This is the type of formulas used when the focus is in motion planning problems over a finite time horizon.
\subsection{LTL Semantics for Motion Trajectories}

The robot workspace $\mathcal{W}$ contains different types of rigid bodies and a set of propositional regions $P=\{P_1\dots P_n\}$ corresponding to the propositions $\{\pi_1 \dots \pi_n\}$. The 
part 
of the workspace that is accessible by the robot is represented as $\mathcal{W}_\textrm{acc}=\mathcal{W} \setminus \mathcal{W}_\textrm{fixed}$, where $\mathcal{W}_\textrm{fixed}$ is the part of the 
workspace occupied by $\mathcal{B}_\textrm{fixed}$. Propositional regions are associated with accessible part of the 
workspace $P \in \mathcal{W}_\textrm{acc}$. A proposition $\pi_0$ is associated with a propositional region such that $P_0=\mathcal{W}_\textrm{acc} \setminus \cup_{i=1}^n P_i$. A function 
$\mathcal{G}:\mathcal{W}_\textrm{acc}\rightarrow \Pi$ maps each point of the workspace over the propositional region.

The discrete trace of a trajectory \textit{Traj} is  defined as the sequence of propositional regions that are traversed by \textit{Traj} and is represented as $tr$(\textit{Traj}). A propositional 
region 
$P_i$ is said to be traversed iff $\mathcal{G}(\textrm{\textit{Traj}}(t))=\pi_i$ for some $0\leq t \leq T$. A motion trajectory \textit{Traj} satisfies $\phi$ iff $tr$(\textit{Traj}) $\models 
\phi$. 
\subsection{Problem Statement}
Let a motion planning problem be considered as the tuple $ \langle E_\textrm{init},  \textrm{\sf \small PROPAGATOR}, \textrm{\sf \small VALIDITYCHECKER}, \mathcal{K}, \kappa, \textrm{ \sf \small 
INFERENCE}, \Pi \\ ,\phi \rangle $ for robot $\mathcal{R}$.
The goal is expressed in terms of an LTL formula~$\phi$, defined over a set of atomic propositions 
$\Pi$ that describe  regions of the workspace that the robot must either visit or avoid.
The problem is to evaluate (by taking into account $E_{init}, \mathcal{K}, \Pi,\phi$) whether the robot can satisfy the formula, if not, whether the formula can be simplified. If the formula (or 
possibly simplified formula) is feasible then the problem is to  find a sequence of control inputs such 
that the resulting robot trajectory \textit{Traj} is satisfied $tr($\textit{Traj}$)\models \phi$ and is 
dynamically feasible, avoids the collision with fixed bodies and, if necessary, pushes 
movable objects away for clearing the regions. 

\begin{figure*}[t]
\begin{center}
   \includegraphics[width=0.85\textwidth]{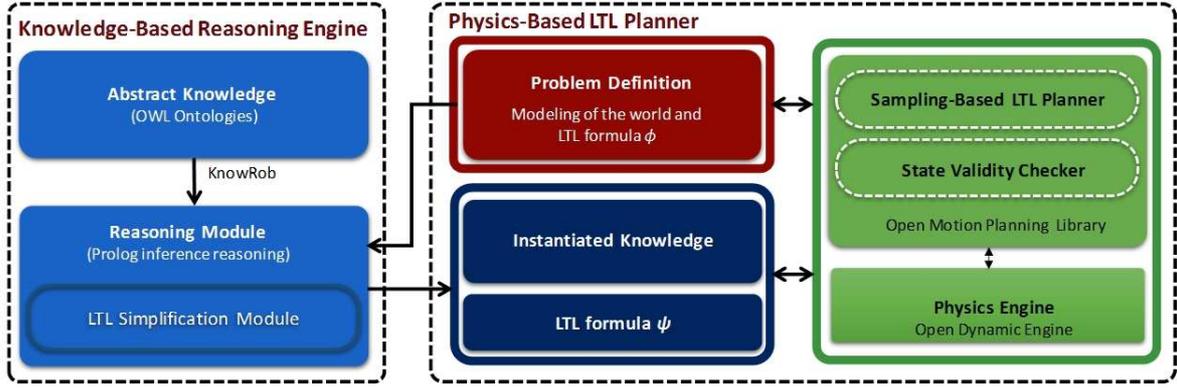}
   \caption{Framework for Physics-based LTL motion planning.}\label{fig:framework}
\end{center}
\end{figure*}

\section{Physic-based LTL Motion Planning}\label{sec:approach}
\subsection{Framework}

To solve the above stated problem, a physics-based LTL motion planning approach is proposed that makes
 use of ontologies and high-level reasoning, and that considers physics-based motion propagation.
 The schematic representation of the solution framework is depicted in Fig.~\ref{fig:framework}. It consists of two main modules: the knowledge-based reasoning engine and the physics-based LTL 
planner.  The former is responsible for defining the manipulation constraints, the feasibility evaluation and the possible simplification of the LTL formula, whereas the latter computes the motion 
plan that satisfies the temporal goals.

The knowledge-based reasoning engine contains the abstract knowledge $\mathcal{K}$, and performs a prolog-based reasoning over $\mathcal{K}$ to define the types of the rigid bodies and the associated 
manipulation constraints (encoded in the instantiated knowledge) using the kinematic and dynamic properties of the robot and the bodies. Furthermore, it is responsible for the feasibility evaluation 
and the possible simplification of the LTL formula (explained in Sec.~\ref{sec:reasoning}). It is important to note that the feasibility evaluation and simplification is different from the validity 
checking of the LTL formulas that is performed using model-checking techniques.

 The physics-based LTL planner stores the problem definition (the world modeling along with the initial LTL formula~$\phi$ that defines the temporal goals),  the instantiated knowledge~$\kappa$ and 
the LTL formula~$\psi$
 inferred by the reasoning module ($\psi$ contains a simplified version of~$\phi$, when possible). The automaton will be constructed for $\psi$, unlike other sampling-based LTL planners that always 
plan for $\phi$. The sampling-based LTL motion planner is responsible for generating the discrete plan (computed over the product space of the decomposed workspace and generated automaton for~$\psi$) 
and its execution at low-level (continuous motion planning level) to determine the control sequence in such a way that the resultant trajectory satisfies~$\psi$.
  The state propagator makes use of the physics engine for the propagation, and the newly generated states are evaluated by the state validity checker. Different from standard sampling-based LTL 
planners, the proposed validity checker takes into account the current state of the environment and evaluates it based on the instantiated knowledge that is valid for that particular state. 

\subsection{Reasoning Process}\label{sec:reasoning}
The aim of knowledge-based reasoning is to provide autonomy to the robots for performing complex tasks. We use the reasoning process, on one hand, to generate the instantiated knowledge~$\kappa$  and, 
on the other hand, to evaluate the feasibility of the LTL formula and its potential simplification. 

For the generation of $\kappa$ a prolog-based reasoning is  employed that reads the abstract knowledge 
in order to classify the objects into different types, together with their manipulation constraints. This process is performed by evaluating physical properties of the objects and the kinodynamic 
properties of the robot in a similar way as performed in~\cite{Muhayyuddin2015, gillani2016}. At each instant of time $\kappa$ is updated using \textrm{\sf \small INFERENCE} function that takes into 
account the previous state of $\kappa$ and current state of the environment and 
generate new state of $\kappa$ \hbox{\textrm{\sf \small 
INFERENCE}: $ E_{i+1} \times \kappa_i\rightarrow \kappa_{i+1}$}. For instance, if after a propagation step one manipulation region is occupied with another object, the motion constraints of the first 
object are updated accordingly 
to the new situation.

\begin{algorithm}[t]
\begin{algorithmic}[1]
{\small \sf
\REQUIRE List $L$ with nonvalid proposition(s), LTL formula $\phi$\\
\ENSURE LTL formula $\phi$\\
\IF {$L.op \ni \vee$}
\RETURN {$\phi \backslash L$}
\ELSE
\IF {$L.parent=\phi$}
\RETURN {\sf \footnotesize NULL}
\ELSE
\STATE  Simplify($L.parent,\phi$)
\ENDIF
\ENDIF
}
\end{algorithmic}
\caption{Simplify}\label{alg-simplify}
\end{algorithm}

\begin{algorithm}[t]
\begin{algorithmic}[1]
{\small \sf
\REQUIRE LTL formula $\phi$,  Set of propositions $\Pi$, Knowledge $\mathcal{K}$\\
\ENSURE $\phi$ or {\small \sf NULL}
\STATE $\mathcal{M}_p\leftarrow$ $\mathcal{F}(\mathcal{K}, \Pi)$
\IF {$\mathcal{M}_p = {\sf \footnotesize NULL}$}
\RETURN {$\phi$}
\ENDIF
\FORALL{$\pi\in \mathcal{M}_p$}
\STATE  $\phi\leftarrow$Simplify($\pi.L,\phi$)
\IF {$\phi = {\sf \footnotesize NULL}$}
\RETURN {\small \sf NULL}
\ENDIF
\ENDFOR
\RETURN {$\phi$}
}
\end{algorithmic}
\caption{Evaluate}\label{alg-evaluate}
\end{algorithm}

\begin{algorithm}[t]
\begin{algorithmic}[1]
{\small \sf
\REQUIRE Initial state $E_{\textrm{init}}$, $\Pi$, LTL formula $\phi$, Threshold $T_{max}$\\
\ENSURE A continuous path that satisfies $\phi$.\\
\STATE $\mathcal{K} \leftarrow$ OntologyFormulation($E_{init}$)
\STATE $\kappa_0 \leftarrow $InstantiatedKnowledgeInference($\mathcal{K}$)
\STATE $\psi \leftarrow$ Evaluate($\mathcal{K},\Pi, \phi$)
\IF {$\psi$ = \small \sf NULL}
\RETURN {\small \sf NULL}
\ELSE
\STATE $\mathcal{T}\leftarrow$InitializeTree($E_{init}$)

\STATE  $\mathcal{A}_{\psi} \leftarrow$ComputeAutomaton($\psi$)
\STATE $\mathcal{D} \leftarrow$ ComputeDecomposition() ; $j=0$
\WHILE {$t<T_{max}$}
\STATE $\rho \leftarrow$ DiscretePlanning($\mathcal{A}_{\psi}$,$\mathcal{D}$ )
\STATE  $v \leftarrow$ SelectHighLevelState($\rho$)
\STATE $\{u,n\}$ $\leftarrow$ SampleControlAndSteps($v$)
 \FOR{$i=0$ \TO $n$ }
\STATE $\mibf{E}_{\textrm{new}}\leftarrow$ PROPAGATOR($E,u$,$\Delta t$)
\IF{! VALIDITYCHECKER($E_{\textrm{new}}$,$\kappa_j$)}
\STATE Break
\ELSE
\STATE  $\kappa_{j+1} \leftarrow$ INFERENCE($E_{new},\kappa_j$); $j=j+1$
\STATE $v_{new}\leftarrow$UpdateHighLevelState($v$)
\STATE $\mathcal{T}\leftarrow$ UpdateTree($E_{new}, u, \Delta t$)
\STATE $z \leftarrow$GetAutomatonState($v_{new}$)
\IF{$z \in \textrm{Accepting state of } \mathcal{A}_{\psi}$}
\RETURN \textit{Traj}$\leftarrow$RetrieveTrajectory($\mathcal{T}$) 
\ENDIF
\ENDIF 
\ENDFOR
\ENDWHILE
\RETURN {\small \sf NULL}
\ENDIF

}
\end{algorithmic}
\caption{Physics-based LTL Motion Planning}\label{algo1}
\end{algorithm}

To evaluate the feasibility  and the possible simplification of the LTL formula, the reasoning process is done as follows:
\begin{itemize}
 \item
Let an LTL formula~$\phi$ be defined over a set of propositions $\Pi=\{\pi_1\dots\pi_n\}$, where each $\pi_i\in\Pi$ is associated with a region $P_i$ of the workspace (that the robot should visit or 
avoid) 
called \textit{propositional region}. A proposition is considered nonvalid if the associated propositional region is not accessible by the robot and valid otherwise.
Let $\mathcal{M}_p$ be the list of nonvalid propositions and
 $\mathcal{F}$ be the function that computes them, i.e. $\mathcal{F}:\mathcal{K}\times\Pi \Rightarrow \mathcal{M}_p$.
If $\mathcal{M}_p = \emptyset$, the formula is feasible and does not require simplification.
\item
Let a formula be considered a list $L$, which can be either an atomic list (defined by a single proposition), or a compound list (defined by the composition of lists using temporal and logic 
operators). 
Then, $L.parent$ will indicate the parent list of a list,  $L.op$ the set of prefix and postfix operators of $L$ within $L.parent$, and $\pi.L$ the innermost list containing the proposition $\pi$ 
(i.e. 
$\pi.L=\{\pi\}$).
\item
Let {\sf \small Simplify($L,\phi$)} be  a recursive function that verifies if $L$ contains disjunction operators and if so returns the formula $\phi$ without $L$, as shown in 
Algorithm~\ref{alg-simplify}.
Then, Algorithm~\ref{alg-evaluate} shows the procedure {\sf \small Evaluate($\phi$)} that checks the feasibility of an LTL formula by using function {\sf \small Simplify($L,\phi$)}.
\end{itemize}

\begin{figure}[t]
\begin{center}
   \includegraphics[width=0.70\columnwidth]{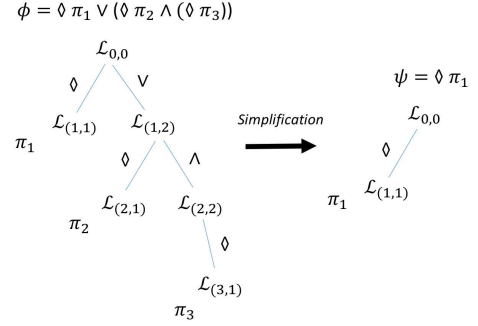}\\
   \caption{Example of the simplification process, where $\phi$ and $\psi$ are the actual and the simplified formulas respectively. Each $\mathcal{L}_{(i,j)}$ is a list with $i$ and $j$ representing 
the depth and the order in the parent list, respectively.}\label{formula}
\end{center}
\end{figure}

As an example, consider the formula \hbox{$\phi=\diamondsuit\pi_1\vee(\diamondsuit\pi_2\wedge(\diamondsuit\pi_3))$} where the proposition  $\pi_3$ is nonvalid. The list associated to $\pi_3$ is 
$\mathcal{L}_{(3,1)}$, as shown in Fig.~\ref{formula}. 
The recursive {\sf \small Simplify} is initially called for  $\mathcal{L}_{(3,1)}$, then for $\mathcal{L}_{(2,2)}$ and $\mathcal{L}_{(1,2)}$, when the $\vee$ operator is found. At this moment the 
formula is 
simplified by deleting $\mathcal{L}_{(1,2)}$ and the simplified formula results $\psi=\Diamond\pi_1$.

The proposed reasoning process works for the syntactically co-safe LTL formulas.
The Open Motion Planning Library (OMPL)~\citep{sucan201}, a C++ based tool for sampling-based motion planning, provides the implementation of the sampling-based LTL motion planner presented 
in~\cite{bhatia2010}. It supports the temporal goals defined using syntactically co-safe LTL formulas such as  $\phi= \diamondsuit \pi_1 \wedge \dots \wedge \diamondsuit $ (visit all the 
regions 
$\pi_1\dots\pi_n$ in any order),  $\phi = \diamondsuit (\pi_1 \wedge \diamondsuit ( \pi_2 \wedge \diamondsuit( \ldots \wedge \diamondsuit \pi_n ) ) )$ (visit all the regions in an ordered way), 
$\phi=\diamondsuit\pi_1\vee \dots \vee\diamondsuit\pi_n$ (visit 
any of the region from $\pi_1\dots\pi_n$), and $\phi=(\diamondsuit\pi_1\vee \diamondsuit\pi_2)\wedge \neg \pi_3$ (visit $\pi_1$ or $\pi_2$ and not visit $\pi_3$). The reasoning process 
works over the similar types of formula with similar grammar.

\subsection{Planning Process}
The planning process is explained in Algorithm \ref{algo1}, that takes as inputs the initial state, the set of propositions $\Pi$, the temporal goal (defined in terms of an LTL formula $\phi$ over 
$\Pi$), and the maximum allowed planning time $T_{max}$. As output it returns a continuous path (as a sequence of controls and durations) that satisfies $\phi$.

The \textit{OntologyFormulation} function 
defines the abstract knowledge~$\mathcal{K}$ about the world by defining the types of the objects (such as fixed or manipulatable), and their manipulation constraints in terms of 
\textit{mRegions}(Sec.\ref{subsec:modeling}).
\textit{InstantiatedKnowledgeInference} fills the initial state of the instantiated knowledge as explained in Sec. \ref{sec:reasoning}.
To determine the feasibility of~$\phi$, the \textit{Evaluate} function computes the feasibility and performs the possible simplification (if required) as explained in Algorithms~\ref{alg-simplify} 
and~\ref{alg-evaluate}.
The function \textit{InitializeTree} sets the initial state of the tree as the initial state of the environment.

Lines: (8-13) refer to the general steps of the sampling-based LTL motion planning, as done in~\cite{bhatia2010}, for the high level planning and for the updating of the 
states 
(both at low- and high- levels). \textit{ComputeAutomaton} function computes the Automaton $\mathcal{A}_\psi$ for the formula~$\psi$, \hbox{\textit{ComputeDecomposition}} performs the triangular 
decomposition of the workspace not occupied by fixed obstacles, in a way that preserves the propositional regions. As a difference with~\cite{bhatia2010} we only exclude from the decomposition the 
workspace occupied by fixed obstacles, i.e. the non-fixed bodies are simply ignored while decomposing the workspace. Function \textit{DiscretePlanning} constructs the discrete plan over the product 
space of the decomposition and $\mathcal{A}_\psi$, and \hbox{\textit{SelectHighLevelState}} selects the high-level state to be explored.
At low level, \hbox{\textit{SampleControlAndSteps}} function samples the controls (that could be a vector of  applied forces, joints torques, or velocities), and the number of steps (that refer to 
the number of times  that the sampled controls will repeatedly applied for a duration~$\Delta t$).

The function \hbox{\sf \small PROPAGATOR} applies the sampled controls for $\Delta t$ time on the robot 
and generates new state of the environment $E_{new}$ using the dynamics engine that allows to 
handle all the kinodynamic and physics-based constraints. \hbox{\sf \small VALIDITYCHECKER}  evaluates the newly generated state of the environment, based on the instantiated knowledge $\kappa$.   
$E_{\textrm{new}}$ will be accepted if it satisfies all the constraints  (such as temporal constraints, kinodynamic and physics-based constraints) that are imposed by $\kappa$ and discarded 
otherwise. 

The \hbox{\sf \small INFERENCE} function updates the instantiated knowledge with the manipulation constraints that are valid for 
$E_{new}$.
\textit{UpdateHighLevelState} updates the high-level state based on the result of the low-level state and \hbox{\textit{UpdateTree}} updates the tree-data structure. The 
\textit{GetAutomatonState} function will determine the state of the automaton, if it is the accepting state of $\mathcal{A}_{\psi}$, the \textit{RetrieveTrajectory} function returns the \textit{Traj} 
that is a continues trajectory for the robot such that $tr(\textrm{\textit{Traj}}) \models \phi$.

\section{Results and Discussion}\label{sec:rnd}

The physics-based LTL motion planning framework (depicted in Fig.~\ref{fig:framework}) is implemented within \textit{The Kautham Project}~\citep{Rosell2014} that is a motion planning framework that 
mainly uses planers from the OMPL~\citep{sucan201}. For the current proposal a variant of the LTL has been implemented, and the Open Dynamic Engine has been used for the state propagation. The 
abstract 
knowledge $\mathcal{K}$ is implemented (in term of OWL ontologies) using the \textit{Prot\'eg\'e} editor.  Instantiated knowledge is defined by applying the prolog-based reasoning 
process, it uses predicates (functions) defined in  Knowrob~\citep{knowrob2009} (a knowledge processing framework for robots) to access the information from $\mathcal{K}$. The communication between 
the modules is performed using ROS~\citep{ros2009}.

\begin{figure}[t]
\begin{center}
   \includegraphics[width=\columnwidth]{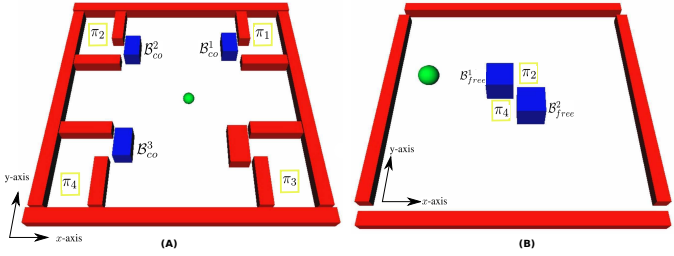}\\
   \caption{Example scenarios A: the goal is to visit the propositional regions $\pi_1, \cdots, \pi_4$, being the access to them obstructed by the blue boxes. B: visit all 
propositional 
regions in an 
ordered 
way, being two of them occupied by bodies. Video: https://sir.upc.edu/projects/kautham/videos/IFAC2017.mp4
}\label{sim:LTL}
\end{center}
\end{figure}

The simulation setup consists of a robot (green sphere),  constraint-oriented movable bodies (blue cubes), and fixed bodies (red walls). There are two scenarios presented in 
Fig.~\ref{sim:LTL}.
In the first scenario, it is assumed that all the propositional regions are surrounded by the objects that could be fixed or movable and no collision free trajectory exist to visit each region. It 
could be considered as different rooms that robot has to visit but in order to enter each room, the robot has to interact with the freely-movable body blocking the entrance. The propositional 
regions 
$\{\pi_1\dots\pi_4\}$, are shown as yellow rectangles. The temporal goal is defined by the LTL formula 
\hbox{$\phi = \diamondsuit (\pi_1 \wedge \diamondsuit ( \pi_2 \wedge \diamondsuit(\pi_3 \vee \pi_4) ) )$} that is visit $\pi_1$, $\pi_2$ and then $\pi_3$ or $\pi_4$
 The region associated with $\pi_3$ is surrounded with $\mathcal{B}_{fixed}$ and therefore, the reasoning process marks $\pi_3$ as invalid (it is not accessible by the robot). Since it has 
disjunction relation with the other propositions, the simplification process will simplify the formula to \hbox{$\psi = \diamondsuit (\pi_1 \wedge \diamondsuit ( \pi_2 \wedge \diamondsuit( 
\pi_4) ) )$}. Since the length of the first body is greater than the entrance, at its current location its manipulation region along the $x$-axis of the world frame is 
occupied with the 
walls. Therefore, the reasoning process will change the status of the body from freely-movable to constraint-oriented movable, and only allow the robot to push it along the y-axis. If after pushing 
the body, all the manipulation regions become free, the \hbox{\sf \small INFERENCE} function will change the 
type of $\mathcal{B}_{co}^1$ to $\mathcal{B}_{free}^1$. The similar process is applied for $\mathcal{B}_{co}^2$ and $\mathcal{B}_{co}^3$.

The temporal goal for the second scenario is described as \hbox{$\phi = \diamondsuit (\pi_1 \wedge \diamondsuit ( \pi_3 \wedge \diamondsuit( \pi_2 \wedge \diamondsuit( \pi_4 ) ) ) )$}. That is, visit 
$\pi_1, \pi_3, \pi_2$ and $\pi_4$ consecutively. The propositional regions associated to $\pi_1$ and $\pi_3$ are occupied by $\mathcal{B}_{free}^1$ and $\mathcal{B}_{free}^2$ respectively. 
Therefore, in order to visit $\pi_1$ (without prior being on $\pi_2$ or $\pi_4$), the robot must push $\mathcal{B}_{free}^1$ along the $x$-axis or $-y$-axis of the world frame. If it pushes 
$\mathcal{B}_{free}^1$ along the $x$-axis then it 
ends occupying the manipulation 
region of 
$\mathcal{B}_{free}^2$ that is along $y$-axis and hence the reasoning process will deactivate the \textit{mRegion} along the $-y$-axis  and change the status of the body to constraint-oriented 
movable. The same reasoning process is repeated for the second body, i.e.  The \hbox{INFERENCE} function will update the types of these bodies to $\mathcal{B}_{co}^1$ and $\mathcal{B}_{co}^2$. To 
visit 
$\pi_3$, 
$\mathcal{B}_{co}^2$ can only be pushed along the $-x$-axis. After visiting $\pi_3$ the types of the bodies will be restored to 
$\mathcal{B}_{free}$ and the task can continue.

These two examples show that the proposed approach is able, on the one hand, to deal with movable objects that may be obstructing the solution path (changing if necessary the way the robot has to 
interact with them) and, on the other hand, is able to simplify a 
formula if part of it is non-feasible.

We tested both scenarios with and without instantiated knowledge. The simulation was performed on
  an Intel Core i7-4500U 1.80GHz CPU with 16 GB memory.  For the first scene, the success rate  of simple physics-based LTL planner was 30\% for 10 runs (maximum allowed time was 300 seconds) whereas 
the success rate of the  proposed approach was 80\%. The simple physics-based LTL approach has an average planning time of 230 seconds. In contrast, the proposed approach computes the solution in 
46.4 
seconds (average of 10 runs).

For the second scenario, the success rate of the proposed approach and the simple physics-based approach were 100\%. But, in the case of the proposed approach the quality of the solution was 
better, it avoids the unnecessary interactions between the robot and the objects and move the objects only when it is necessary. Regarding 
planning time, the proposed approach computes the solution in 2.1 seconds (average of 10 runs) and simple physics-based planning approach takes 23.8 seconds. 

\section{Conclusions}\label{sec:concl}
This paper has proposed the integration of LTL planning within the framework of ontological physics-based motion planning in order to provide robustness and autonomy for handling complex temporal 
goals in a realistic way. Moreover, a simplification process of the LTL formula is proposed, according to the validity or not of the goals to be satisfied and the logical operators involved. The 
proposed approach has been validated using simulation examples in which some of the propositional regions are occupied with objects or the way to the propositional region is blocked with objects that 
the robot has to push away, if possible, in order to visit the 
regions. The results shows that the integration of knowledge makes the planner more efficient and enhance the quality of the solution.
\balance
\bibliography{References} 
\end{document}